\documentclass[letterpaper,10pt,conference]{ieeeconf}
\IEEEoverridecommandlockouts
\usepackage{cite,amsmath,amssymb,graphicx,booktabs,array,url,xcolor,microtype}
\usepackage{tikz}
\usepackage{float}
\usepackage{capt-of}
\newcommand{\method}{FootQuery}
\newcommand{\R}{\mathbb{R}}
\newcommand{\sg}{\operatorname{sg}}
\title{FootQuery: Future-Touchdown-Guided Retrieval from Depth History for Perceptive Humanoid Locomotion}
\author{Tao Dong$^{1,*}$, Jia Yu$^{2,*}$, Yuxuan Fan$^{3}$, Linna Zhao$^{2}$, Jiaqi Gong$^{2}$, Andong Yang$^{4}$, Chao Gao$^{1,\dagger}$, Guyue Zhou$^{1}$%
\thanks{$^{*}$Equal contribution. $^{\dagger}$Corresponding author.}%
\thanks{$^{1}$Institute for AI Industry Research (AIR), Tsinghua University, Beijing, China.}%
\thanks{$^{2}$University of Science and Technology Beijing, Beijing, China.}%
\thanks{$^{3}$Nanyang Technological University, Singapore.}%
\thanks{$^{4}$Department of Electrical Engineering. Tsinghua University, Beijing, China.}%
}
\IEEEaftertitletext{%
\begin{minipage}{\textwidth}
\centering
\includegraphics[width=\textwidth]{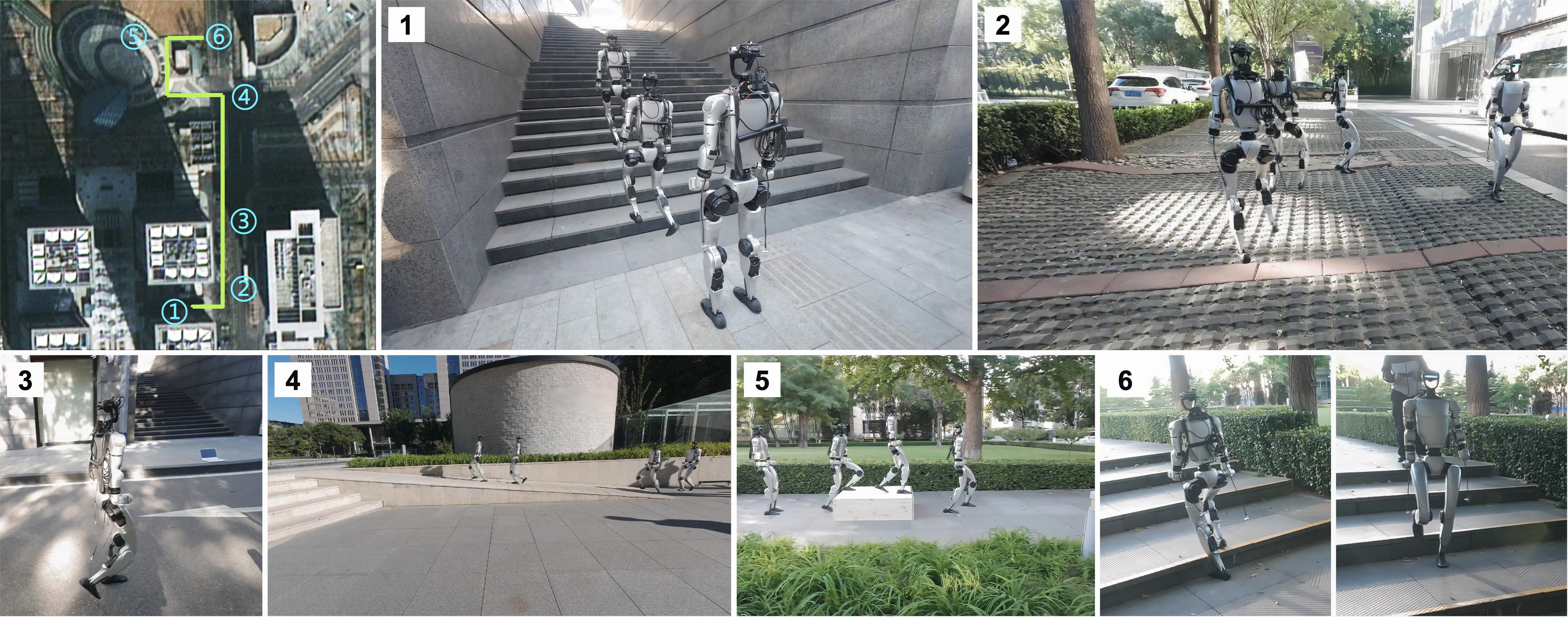}
\captionof{figure}{\textbf{FootQuery on outdoor terrains.} The Unitree G1 traverses stairs, paved surfaces, and walkways. The aerial view locates scenes 1--6 along the route; overlaid robot poses show successive motion stages.}
\label{fig:outdoor-teaser}
\end{minipage}\vspace{0.8\baselineskip}}
\begin{document}
\maketitle

\begin{abstract}
Humanoid locomotion over complex terrain requires anticipating footholds that may no longer be visible at touchdown. Limited camera coverage and self-occlusion make it necessary to retrieve relevant terrain information from earlier observations. We present FootQuery, a perceptive locomotion framework that queries depth history using each foot's predicted next touchdown. The policy predicts touchdown locations and uncertainty from proprioception and uses these distributions, together with per-foot features, to query sparsely sampled historical depth frames. During training, realized contacts are projected into historical images to supervise retrieval at the regions where those contacts were visible. The retrieved per-foot features are fused with global visual memory to generate control actions. A progressive force-assistance curriculum supports early exploration, while event-consistent tread-midline shaping encourages coordinated stair contacts. Deployment requires only proprioception and onboard depth images. In simulation, the complete framework outperforms its component ablations on the most challenging tested stairs, gaps, and platforms. Real-world experiments on a Unitree G1 demonstrate continuous traversal with a single policy across outdoor stairs and indoor routes combining stair ascent and descent, platforms, and gaps. These results support organizing visual history around anticipated contacts for perceptive humanoid locomotion. Project website: https://beiketaoerge.github.io/footquery/.
\end{abstract}

\raggedbottom
\section{Introduction}

Traversing stairs, gaps, and sparse support surfaces requires a humanoid robot to anticipate where its feet will land and adjust its motion before contact. Reinforcement learning has enabled quadrupeds to negotiate such obstacles using onboard vision\cite{Aga22,Mik22,Zhuang23Parkour}, with subsequent work extending perceptive locomotion to bipeds and humanoids\cite{Dua23,Zhuang25Humanoid,Wan25,Zhu26}. Vision provides information about the height, position, and extent of upcoming support surfaces that proprioception alone cannot supply\cite{Yan21}. Yet this information may disappear from the current view as the robot advances or its legs occlude the camera, a visual constraint that in manipulation has been addressed by fusing complementary sensing\cite{Jia25GelFusion}. A region observed earlier can therefore remain relevant to a future touchdown after it is no longer visible\cite{Aga22,Luo26,Yan23}.

Existing approaches address terrain perception through complementary representations. Local elevation reconstruction and history integration retain previously observed terrain\cite{Yu26START,Luo26,Sun26SOLO}, while proprioceptive state and future reference motions guide the selection of motion-relevant features\cite{He25b,Fu26,Hao26,Par26DELTA,Li26PGMT}. Terrain-aware planners generate desired footholds and trajectories\cite{Gra23,Aco25}; current foot-position encodings, local elevation supervision, and touchdown prediction provide additional structure for policy learning\cite{Hwa26,Li25b,Yu26}. Together, these approaches establish the value of retaining terrain information and relating it to the robot's motion.

For a policy operating on visual history, the remaining challenge is to select the frames and image regions relevant to each foot's next contact. The two feet touch down at different locations and times, so their information needs differ. Current foot positions describe the present configuration, and reference motions specify desired movement\cite{Li26PGMT,Hwa26}. Predicting the next realized touchdown provides a contact-specific condition for retrieval. It also defines a training signal: once a contact occurs, its location can be traced back to the historical images in which it was visible. This supervision directly links terrain selection to future support, complementing history retention and feature selection based on robot state or reference motion\cite{Sun26SOLO,Hao26,Par26DELTA,Li26PGMT}.

\begin{figure}[t]
\centering
\includegraphics[width=\columnwidth]{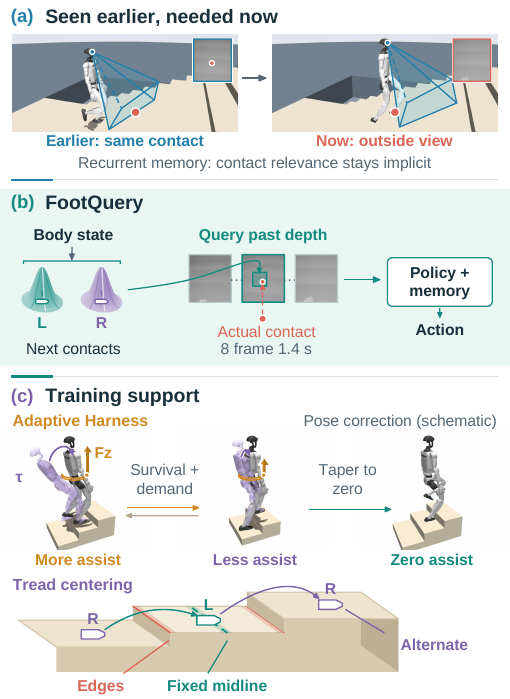}
\caption{FootQuery and its training support. (a) A future contact region leaves the current view. (b) Predicted per-foot touchdown distributions guide depth-history queries; realized contacts supervise retrieval during training. (c) Progressive force assistance and tread-midline rewards support exploration and successive stair contacts. Distributions and read regions are schematic.}
\label{fig:footquery-overview}
\end{figure}

We introduce FootQuery to implement this contact-conditioned retrieval, as illustrated in Fig.~\ref{fig:footquery-overview}. Current and historical proprioception are encoded into per-foot features, from which the policy predicts each foot's next touchdown location and uncertainty. The predicted distributions and per-foot features jointly form queries over sparsely sampled depth history. During training, realized touchdown locations are projected into historical depth images, and the regions where they were visible supervise the attention weights. The policy combines the retrieved per-foot features with global visual memory to generate control actions. At deployment, it requires only proprioception and onboard depth history.

We also introduce two training mechanisms to support exploration and successive stair contacts. A progressive force-assistance curriculum regulates pelvis support using force demand and survival statistics, then withdraws assistance to zero. Event-consistent tread-midline shaping locks a target tread early in swing and rewards progress toward its midline, together with alternating contacts on successive treads. We evaluate the complete framework through simulation comparisons, component ablations, and indoor and outdoor experiments on the Unitree G1. Figure~\ref{fig:outdoor-teaser} shows outdoor traversal on stairs and paved surfaces.

Our contributions are:

\begin{itemize}
\item A foot-specific retrieval mechanism that uses predicted touchdown distributions to query depth history, with direct supervision from historical image regions where realized contacts were visible.

\item A progressive force-assistance curriculum that adapts support using pre-clamp force demand and survival statistics, followed by a monotonic withdrawal to zero.

\item An event-consistent tread-midline shaping objective that fixes the target tread early in swing and encourages centered, alternating contacts on stairs.

\item An evaluation combining simulation comparisons and component ablations with single-policy hardware demonstrations across outdoor stairs and indoor routes containing stairs, platforms, and gaps.
\end{itemize}

\section{Related Work}

\subsection{Visual Perception and Terrain Representation}

Visual locomotion control requires retaining terrain observed during motion and extracting information relevant to action generation. NVM\cite{Yan23} aligns historical 3D features to the current coordinate frame and aggregates them, allowing previously visible terrain to inform control. SOLO\cite{Sun26SOLO} queries sensor histories using map-cell coordinates and learns terrain representations at the corresponding locations through elevation reconstruction. These methods aggregate historical features or reconstruct terrain to make information beyond the current field of view available to the policy.

For terrain feature selection, CReF\cite{Hao26} uses proprioceptive state and estimated velocity to query features from the current depth image, then integrates temporal information through a recurrent network. DELTA\cite{Par26DELTA} adjusts elevation-map sampling based on robot state and local geometry and observes an association between attention and subsequent touchdown locations. In target tracking, DIMM\cite{Zha26DIMM} uses reinforcement learning to adaptively fuse estimates from decoupled motion-model filters under measurement uncertainty. We further focus retrieval on each foot's next touchdown, querying retained depth-history features separately for the two feet. During training, regions where realized touchdowns were visible in historical images supervise per-foot attention weights. Whereas elevation reconstruction targets map-cell heights, our supervision targets historical image regions associated with specific contacts, directly guiding the selection of relevant frames and image regions.

\subsection{Foothold Prediction and Contact Learning}

Footstep planning specifies desired contacts, whereas touchdown prediction estimates contacts resulting from executed motion. MPFC\cite{Aco25} jointly plans foothold regions, touchdown positions, and step durations, continually updating the plan during locomotion. During training, SSR\cite{Yu26} predicts future touchdown locations and uncertainty from privileged state and actions, supervises these predictions with realized future contacts, and uses the predicted distributions to construct support rewards before touchdown. We predict each foot's next touchdown location and uncertainty from current and historical proprioception and use these predictions to condition historical visual queries at deployment. Whereas SSR uses predictions to construct training rewards, our predictor participates in perception during policy execution, guiding the retrieval of historical terrain features around anticipated contacts.

Contact learning also involves foothold rewards and early exploration. BeamDojo\cite{Wan25} evaluates support quality through sole sampling and combines two value networks with two-stage training to learn locomotion on sparse support surfaces. CReF\cite{Hao26} fixes a set of supportable foothold candidates at liftoff and rewards the distance from the realized touchdown to the nearest candidate. For successive stair contacts, we fix a target tread early in swing and reward progress toward its midline and alternating touchdowns, guiding both touchdown placement and contact order on adjacent treads. To support exploration, A2CF\cite{Cao25} learns assistive forces and torques at the pelvis and progressively withdraws assistance through bound adaptation, random masking, and force penalties. We adjust assistance stages using pre-clamp demand and survival statistics, followed by a prescribed withdrawal to zero, transitioning training from assisted exploration to unassisted locomotion.

\section{Method}
\flushbottom
\label{sec:method}
FootQuery is the core policy architecture; force assistance and tread-midline shaping support training only.

\subsection{Problem Formulation}
\label{sec:setting}
We formulate perceptive locomotion as a partially observable Markov decision process and train an asymmetric actor--critic with PPO\cite{Schulman17}.

\noindent\textit{Observations and actions.} For $n$ actuated joints, current proprioception is
\begin{equation}
 o_t=[\omega_t^b,g_t^b,u_t^{\rm cmd},q_t-q^0,\dot q_t-\dot q^0,a_{t-1}],
\end{equation}
comprising body angular velocity, projected gravity, planar velocity commands, joint-state deviations from the default pose, and the previous action. The actor also receives proprioceptive history $H_t$ and depth history $D_t$. The critic receives clean current proprioception and privileged base velocity, terrain heights, and foot-height/contact information, with histories restricted to the actor. Actions $a_t\in\R^n$ specify scaled joint-position offsets; encoder-bias compensation and joint-range clipping produce the applied targets, while a limit penalty acts on unclipped targets.

\noindent\textit{Termination and rewards.} Episodes time out after 20~s with PPO bootstrapping. Excessive tilt, insufficient pelvis height, gap falls, or sustained shin/linkage contact trigger failure. Reaching within 0.45~m of a navigation target with valid orientation records success and refreshes the target within the same episode. Environment rewards combine command tracking, target progress, posture/motion regularization, and support, slip, impact, and limit penalties. When enabled, AMP is mixed with task reward on eligible transitions. Prediction and read losses provide auxiliary supervision.

\subsection{Policy Architecture with FootQuery}
\label{sec:footquery}
\begin{figure}[t]
\centering
\includegraphics[width=\columnwidth]{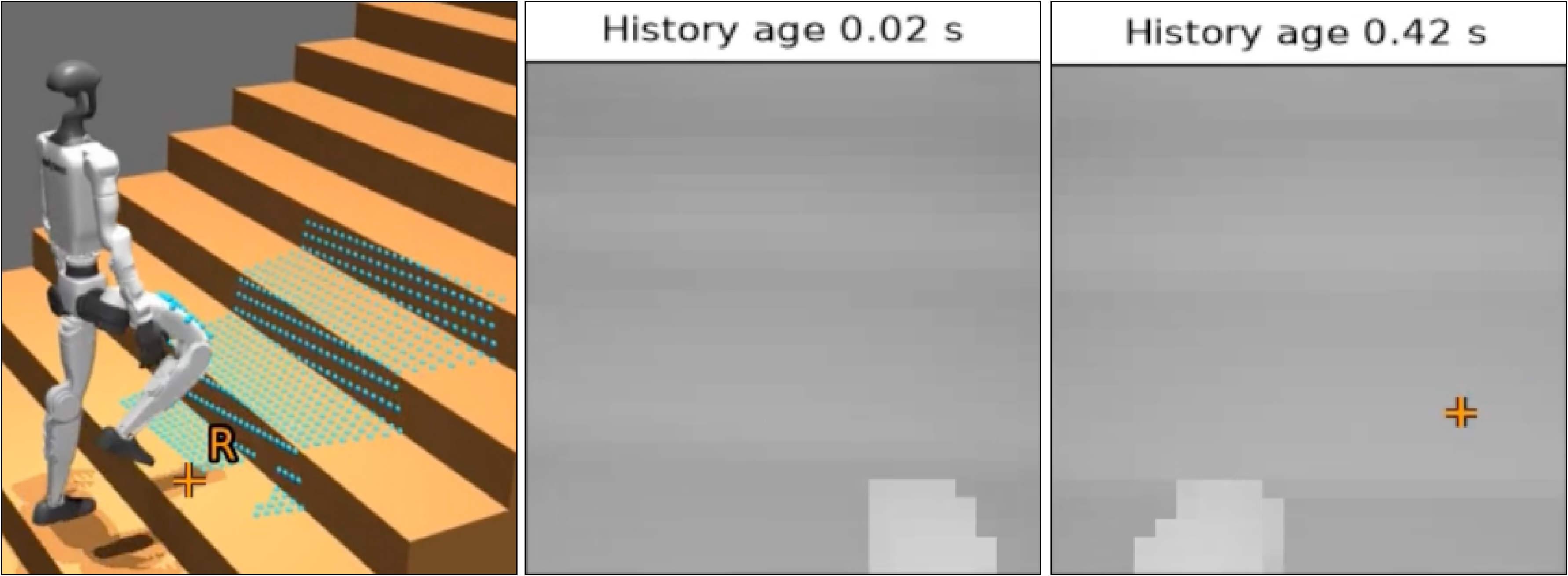}
\caption{Historical visibility of a future contact. Left: cyan points show visible terrain during stair ascent; orange crosses mark the realized future right-foot touchdown. Middle/right: depth frames aged 0.02/0.42~s. The touchdown is outside the latest ROI but visible in the older frame. Contact markers are offline references.}
\label{fig:motivation}
\end{figure}

\begin{figure*}[t]
\centering
\includegraphics[width=0.96\textwidth]{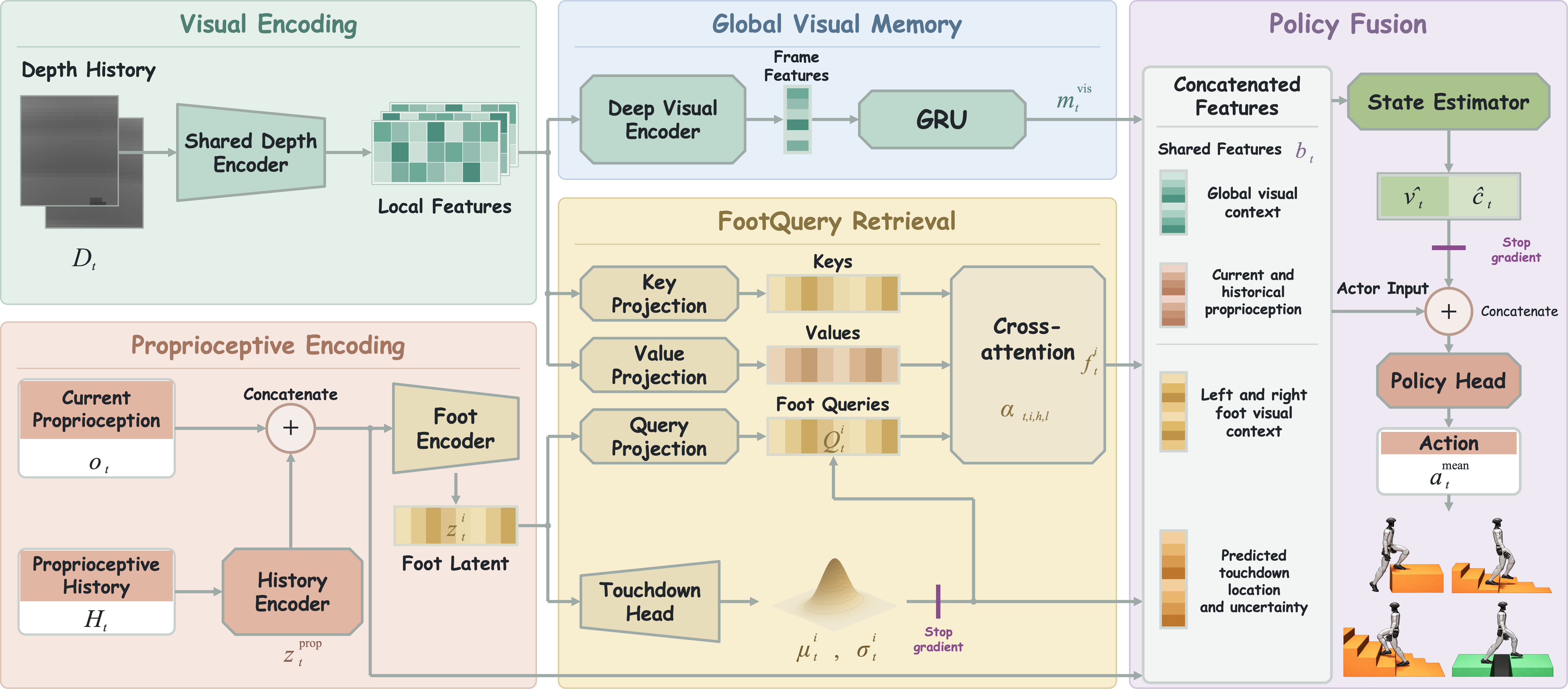}
\caption{FootQuery architecture. Historical depth supplies global memory and local tokens. Proprioceptive foot features and detached touchdown distributions condition left/right queries. State estimation uses global visual and proprioceptive features; the action head additionally receives foot contexts and detached predictions. Purple bars mark stop-gradient.}
\label{fig:architecture}
\end{figure*}

A future support region can leave the camera view or become occluded by the legs before contact. As shown in Fig.~\ref{fig:motivation}, a future touchdown outside the latest ROI was visible 0.42~s earlier. FootQuery uses proprioceptive touchdown predictions to retrieve this historical terrain evidence. Fig.~\ref{fig:architecture} illustrates the visual encoding, per-foot prediction, and policy-fusion paths.

\subsubsection{Historical Visual Encoding}
A shared shallow CNN encodes each depth frame into local features $F_t^{(k)}$. Learned frame and spatial embeddings augment their cells to form 288 historical tokens for key/value projection. Tokens with less than 5\% valid-depth coverage in their receptive fields are masked. Retrieval uses these visual features and embeddings without camera-pose alignment.

Deeper encoding and global average pooling produce frame-level features $g_t^{(k)}$. A GRU\cite{Cho14} processes them from oldest to newest to form global visual memory $m_t^{\rm vis}$, starting from zero for each history window. Local tokens preserve spatial evidence for retrieval, while this window-level memory summarizes the scene.

\subsubsection{Proprioceptive Encoding and Touchdown Prediction}
A History Encoder maps $H_t$ to $z_t^{\rm prop}$. The shared Foot Encoder forms $z_t^i=F_\Theta([o_t,z_t^{\rm prop}],e_i)$ using foot identity $e_i$. A separate Touchdown Head predicts each next contact:
\begin{equation}
 (\mu_t^i,\sigma_t^i)=P_\Theta(z_t^i),\qquad i\in\{L,R\}.
\end{equation}
Here $\Theta$ denotes learnable parameters; $\mu_t^i$ and $\sigma_t^i$ define a diagonal Gaussian in the current pelvis-yaw XY frame. A scaled $\tanh$ bounds the mean by $s_{xy}=(1.0,0.6)$~m, and softplus with clipping bounds each standard deviation to $[0.03,0.60]$~m. The predictor uses proprioception alone to estimate where the policy's motion will place the foot.

\subsubsection{Contact-Conditioned Retrieval and Action Generation}
Query Projection and Touchdown Head are independent heads sharing $z_t^i$. Their outputs are explicitly connected through a learned distribution embedding:
\begin{equation}
 Q_t^i=W_Qz_t^i+\phi\!\left(\sg[\mu_t^i/s_{xy},\sigma_t^i/s_{xy}]\right),
\label{eq:query}
\end{equation}
where $\sg$ denotes stop-gradient. Four-head cross-attention\cite{Vaswani17} matches the query to historical keys and aggregates values into a 64-dimensional foot context $f_t^i$. Its weights $\alpha_{t,i,h,l}$, for head $h$ and token $l$, normalize over valid tokens; all-invalid histories return zero context. The distribution embedding conditions retrieval over the token history.

Policy fusion forms $b_t=[o_t,z_t^{\rm prop},m_t^{\rm vis}]$. The State Estimator infers base linear velocity and contact probabilities as $(\widehat v_t,\widehat c_t)=S_\Theta(b_t)$. The action MLP additionally receives foot contexts, touchdown distributions, and detached state estimates:
\begin{equation}
\begin{aligned}
 f_t^{\rm act}&=[b_t,f_t^L,f_t^R,\sg[\mu_t^L,\sigma_t^L,\mu_t^R,\sigma_t^R]],\\
 a_t^{\rm mean}&=A_\Theta([f_t^{\rm act},\sg[\widehat v_t,\widehat c_t]]).
\end{aligned}
\label{eq:actor}
\end{equation}

\subsubsection{Joint Supervision and Gradient Flow}
Within each 24-step rollout, we label the first touchdown of the current or next swing, requiring at least three airborne steps from prediction time or subsequent liftoff. Labels stay within the episode and rollout; an invalid first touchdown leaves the label masked. Contacts are expressed in the prediction-time pelvis-yaw frame and clipped to $\pm s_{xy}$ for regression. Mean prediction uses Smooth-$L_1$ loss $\mathcal L_{xy}$ with a 0.05-m threshold, and the Gaussian uses NLL $\mathcal L_{\rm nll}$. Both are averaged with weights $w=0.25+0.75(1-\Delta t_{\rm td}/24)$, where $\Delta t_{\rm td}$ is the lead in control steps.

For read supervision, training-only camera poses project unclipped realized contacts into historical depth frames. In-ROI, valid-depth projections consistent within 0.12~m produce bilinear labels $G_{t,i,l}$ on the CNN token grid using the same flip/crop coordinates. Each visible frame is normalized separately. For the visible-target set $\mathcal G$,
\begin{equation}
 \mathcal L_{\rm read}=\big\langle-\log\max(\epsilon,
 \max_h\sum_l\alpha_{t,i,h,l}G_{t,i,l})\big\rangle_{\mathcal G,w},
\label{eq:read}
\end{equation}
where $\epsilon=10^{-8}$ and brackets denote the weighted mean; empty sets contribute zero loss. The best-head reduction allows a head to specialize in retrieving observed contact evidence.

A single optimizer jointly minimizes
\begin{equation}
 \mathcal L=\mathcal L_{\rm PPO}+\mathcal L_{\rm est}+0.5\mathcal L_{xy}+0.05\mathcal L_{\rm nll}+0.01\mathcal L_{\rm read},
\label{eq:joint-loss}
\end{equation}
with $\mathcal L_{\rm est}=0.1\mathcal L_v+0.05\mathcal L_c$ supervising simulated velocity and contact using Smooth-$L_1$ and binary cross-entropy. XY/NLL update the Touchdown Head and shared proprioceptive/foot encoders. The read loss directly updates Q/K, shallow CNN features, embeddings, and the foot encoder; V projection, deeper CNN, GRU, action head, and critic receive no direct read-loss gradient. PPO trains feature and action paths. Detachment blocks direct PPO updates to prediction heads while preserving updates through shared encoders.

\subsection{Progressive Force-Assistance Curriculum}
\label{sec:assist}
\begin{figure}[t]
\centering
\includegraphics[width=0.95\linewidth]{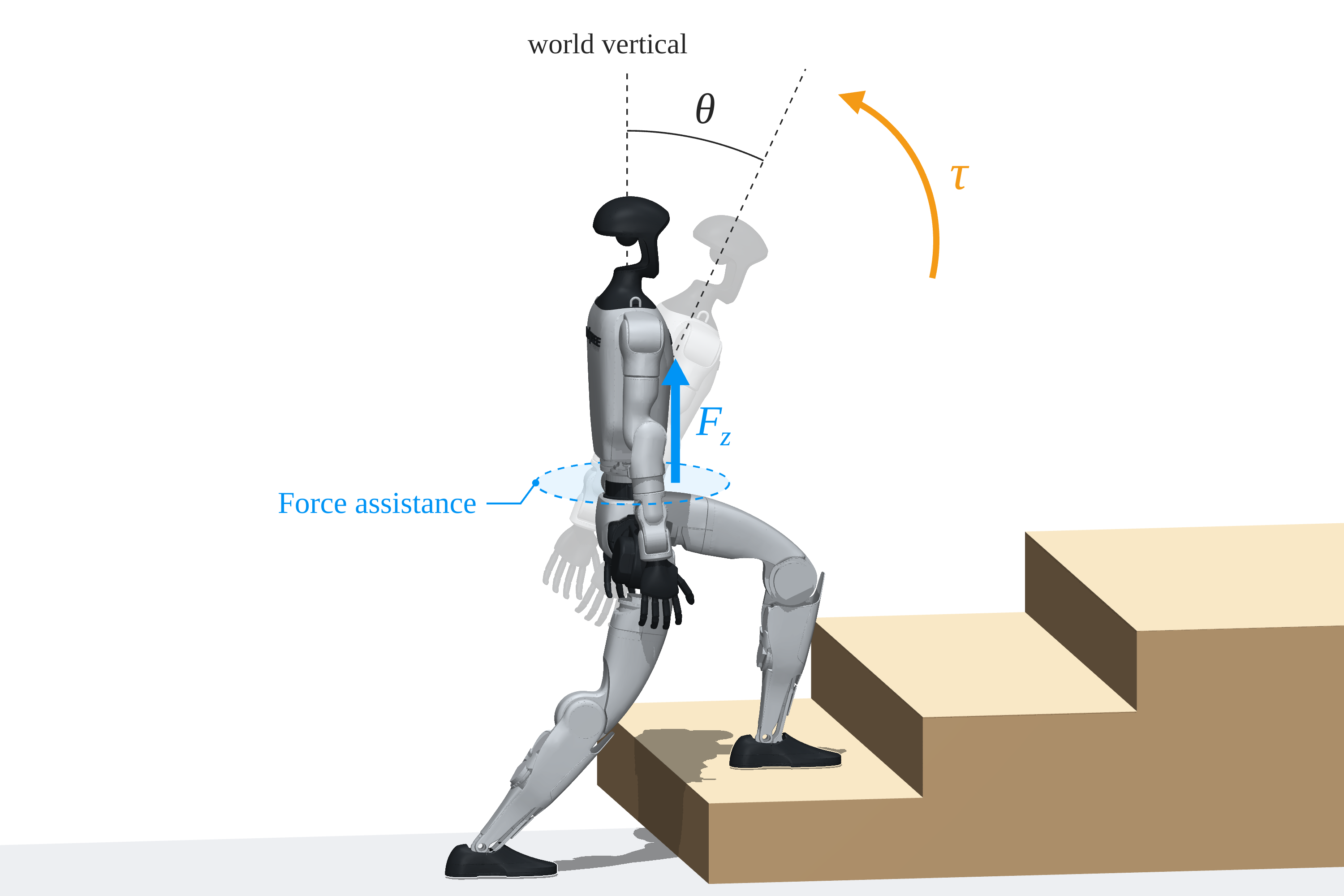}
\caption{Pelvis assistance during stair ascent. Blue: upward support $F_z$. Orange: restoring torque $\tau$. The translucent pose shows tilt $\theta$ relative to world vertical.}
\label{fig:force-assistance}
\end{figure}

Fig.~\ref{fig:force-assistance} illustrates virtual pelvis support: vertical spring--damper force and roll/pitch restoring torques aid early exploration. Per-terrain caps limit both. Before Stage 4, a 32-episode window permits upgrading when survival is at least 0.5 and the P90 of episode-mean pre-clamp force and torque demands each falls below 0.8 times the next cap. Survival below 0.25 permits downgrading. Pre-clamp measurement keeps the demand criterion independent of clipping.

On reaching Stage 4 ($70$~N/$25$~Nm), or at step 72,000 at the latest, continuous withdrawal begins from the current caps:
\begin{equation}
 (F_{\lim},\tau_{\lim})=(F_{\lim}^0,\tau_{\lim}^0)(1-p)^{1.5},
\end{equation}
Here $t_0$ is the withdrawal onset step and $p=\operatorname{clip}((t-t_0)/168000,0,1)$. Stage changes stop during withdrawal; at $p=1$ both caps remain exactly zero. Demand/usage penalties discourage dependence on assistance, while terrain difficulty adapts independently. Evaluation and deployment use zero auxiliary wrench.

\subsection{Tread-Midline Shaping on Stairs}
\label{sec:stair-shaping}
\begin{figure}[t]
\centering
\includegraphics[width=\columnwidth]{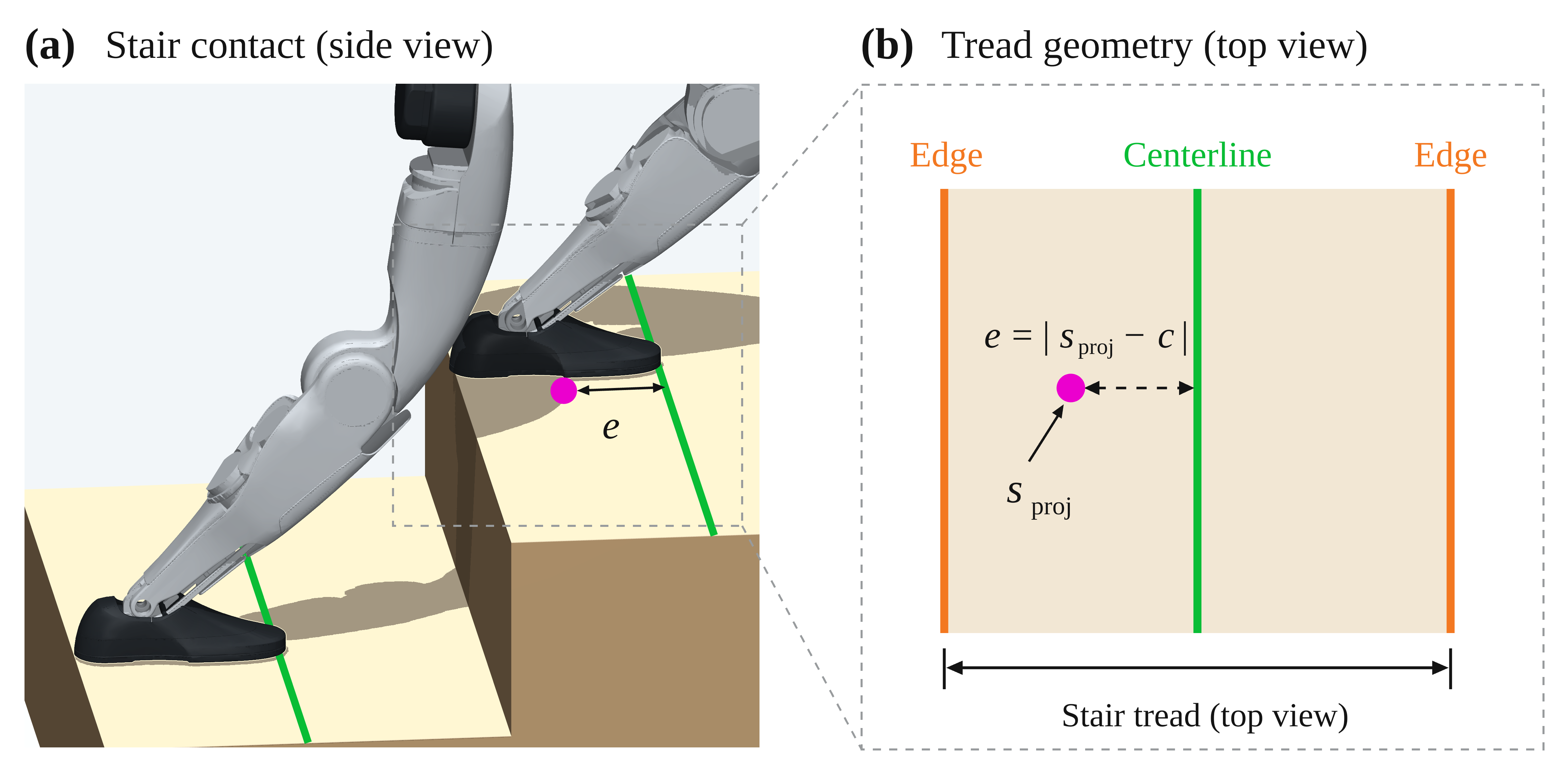}
\caption{Tread-midline geometry. (a) Stair contact side view. (b) Selected tread from above: orange edges, green midline, and magenta projected sole center. Travel-direction error is $e=|s_{\rm proj}-c|=|u_t^{\rm eval}-c|$.}
\label{fig:stair-midline-geometry}
\end{figure}

A training-only pelvis-yaw ray scan detects tread edges and centers. The next target tread is locked in early swing relative to the last committed tread; assigning one foot to it prevents target switching and competing claims. Sole-center progress $u_t^{\rm eval}$ uses short swing lookahead and the measured touchdown position. Its travel-direction error to tread center $c$ in Fig.~\ref{fig:stair-midline-geometry} defines
\begin{equation}
 \bar\varepsilon_t=\operatorname{clip}\!\left(
 \frac{|u_t^{\rm eval}-c|-0.03}{0.13},0,1\right),\quad
 \Phi_t=-\bar\varepsilon_t^2.
\end{equation}
The shaping reward is $\Phi_t-\Phi_{t-1}$, with $-0.25\bar\varepsilon_t^2$ added at valid touchdown. Potential differences reward progress toward the midline; the stored potential resets when the target changes. A sequence reward gives $+1$ for alternating onto the next tread and $-1$ when both feet occupy the same tread. Centering and sequencing have weights 2.0 and 1.0 and apply only on stairs, leaving lateral placement unconstrained.

\begin{figure*}[t]
\centering
\includegraphics[width=0.85\textwidth]{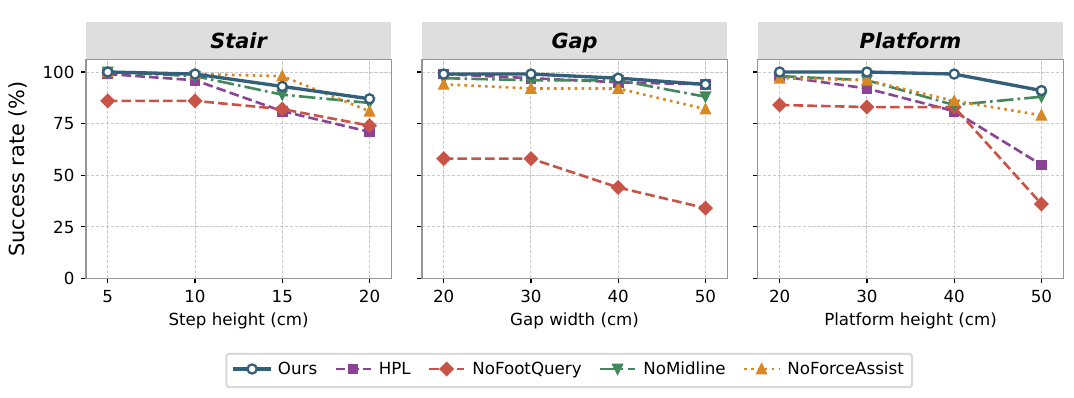}
\caption{Success rates for Ours, HPL, and three component ablations across terrain difficulty. Stair results are not separated into ascent and descent. All panels use the same percentage scale.}
\label{fig:component-ablation}
\end{figure*}

\section{Experiments}
\label{sec:experiments}
\subsection{Experimental Configurations}
\noindent\textbf{Training Environment:} We train policies for the Unitree G1 humanoid in MuJoCo using the mjlab framework on a single NVIDIA RTX 5090 GPU. The main configuration controls 21 joints: 12 leg joints, waist yaw, and eight shoulder/elbow joints; waist roll/pitch and the wrist joints are fixed. PPO collects 24 control steps per rollout. Physics runs at 200~Hz and policy control at 50~Hz. Training terrains include flat ground, rough surfaces, stairs, slopes, raised platforms, and gaps, with terrain difficulty adjusted through curriculum learning.

\noindent\textbf{Depth and Proprioceptive History:} Simulated depth is acquired by ray casting at a nominal 30~Hz with a $36\times60$ source resolution (height $\times$ width). Images are vertically reoriented and cropped to a near-field $27\times24$ ROI before buffering. Eight frames are retained at six-camera-frame intervals, spanning approximately 1.4~s. Proprioception combines the current observation with nine preceding control frames; the historical frames exclude velocity commands. Proprioceptive and depth noise are applied during training.

\noindent\textbf{Inference Interface:} The exported policy receives current proprioception, proprioceptive history, and depth history, and produces 21 actions converted to bounded joint-position targets. Ground-truth touchdowns and historical camera poses are used only for training supervision, not as inference inputs. Force assistance is disabled during evaluation.

\subsection{Locomotion Performance and Component Ablations}
\label{sec:component-ablation}
We compare FootQuery with Humanoid Parkour Learning (HPL)~\cite{Zhuang25Humanoid} and three ablations: NoFootQuery, NoMidline, and NoForceAssist. We also evaluate continuous mixed-terrain traversal in simulation: the same policy links successive locomotion stages as support geometry changes across terrain transitions.

Fig.~\ref{fig:component-ablation} compares the success rates of Ours, HPL, and the ablations. At the largest settings (20-cm stairs, 50-cm gaps and platforms), Ours achieves 87\%, 94\%, and 91\% success, respectively. Relative to NoFootQuery, the gains are 13, 60, and 55 percentage points, with the largest differences on gaps and platforms. HPL matches Ours on the 50-cm gap (94\%), but reaches 71\% on 20-cm stairs and 55\% on the 50-cm platform.

Force assistance has difficulty-dependent effects: NoForceAssist reaches 98\% on 15-cm stairs versus 93\% for Ours, while removing assistance lowers success at the largest stair, gap, and platform settings.

\begin{figure}[t]
\centering
\includegraphics[width=\columnwidth]{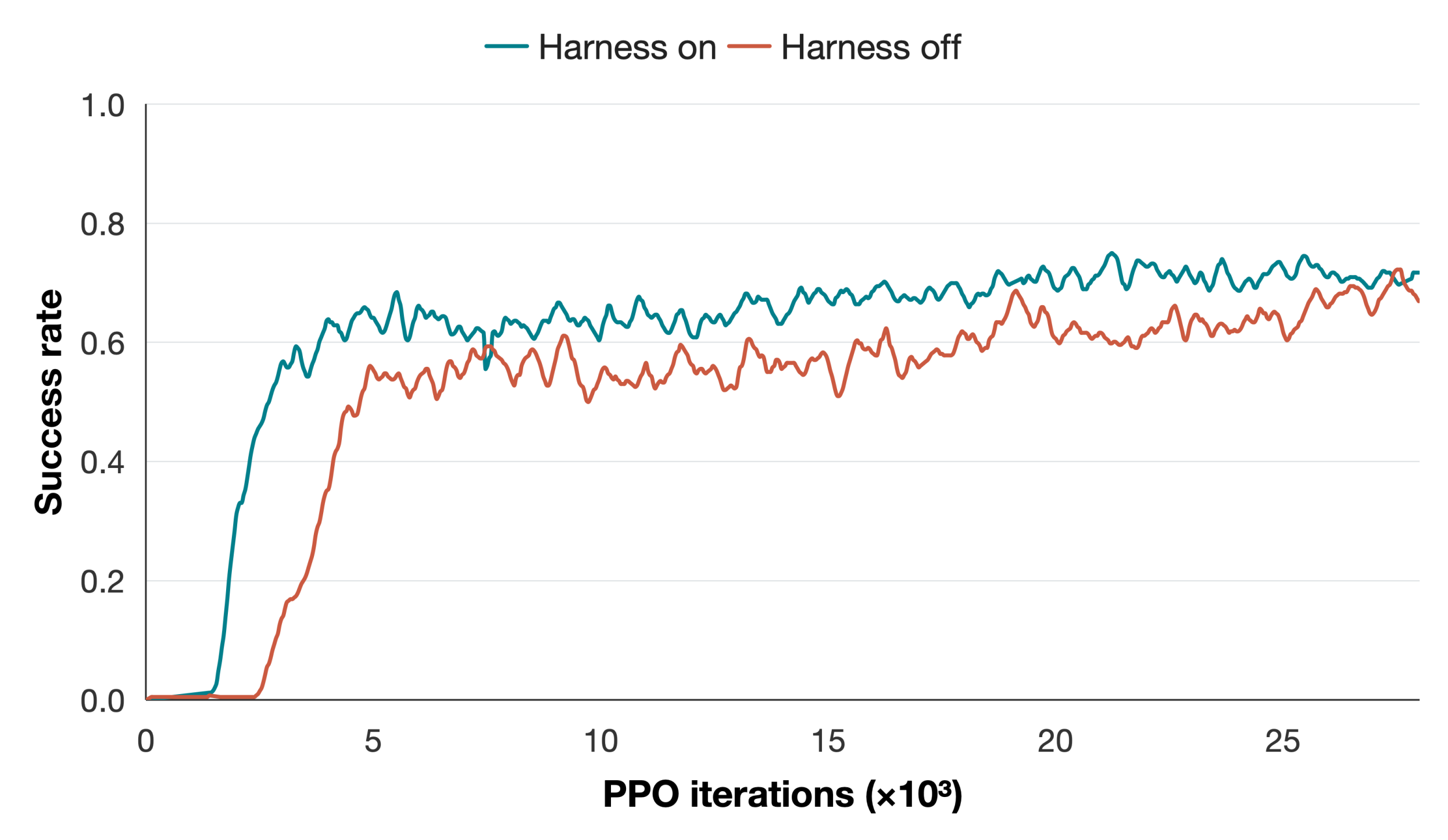}
\caption{All-terrain training success with and without progressive force assistance. One pair of aggregate traces shows success versus PPO iteration.}
\label{fig:force-assistance-training}
\end{figure}

The learning curves in Fig.~\ref{fig:force-assistance-training} reach 0.5 aggregate success roughly 1.5--2 thousand PPO iterations earlier with assistance. The gap narrows late in training. These traces support faster early optimization with assistance. Terrain difficulty follows successful experience, but the curves do not resolve terrain-level transition times.

\begin{figure}[t]
\centering
\includegraphics[width=0.95\linewidth]{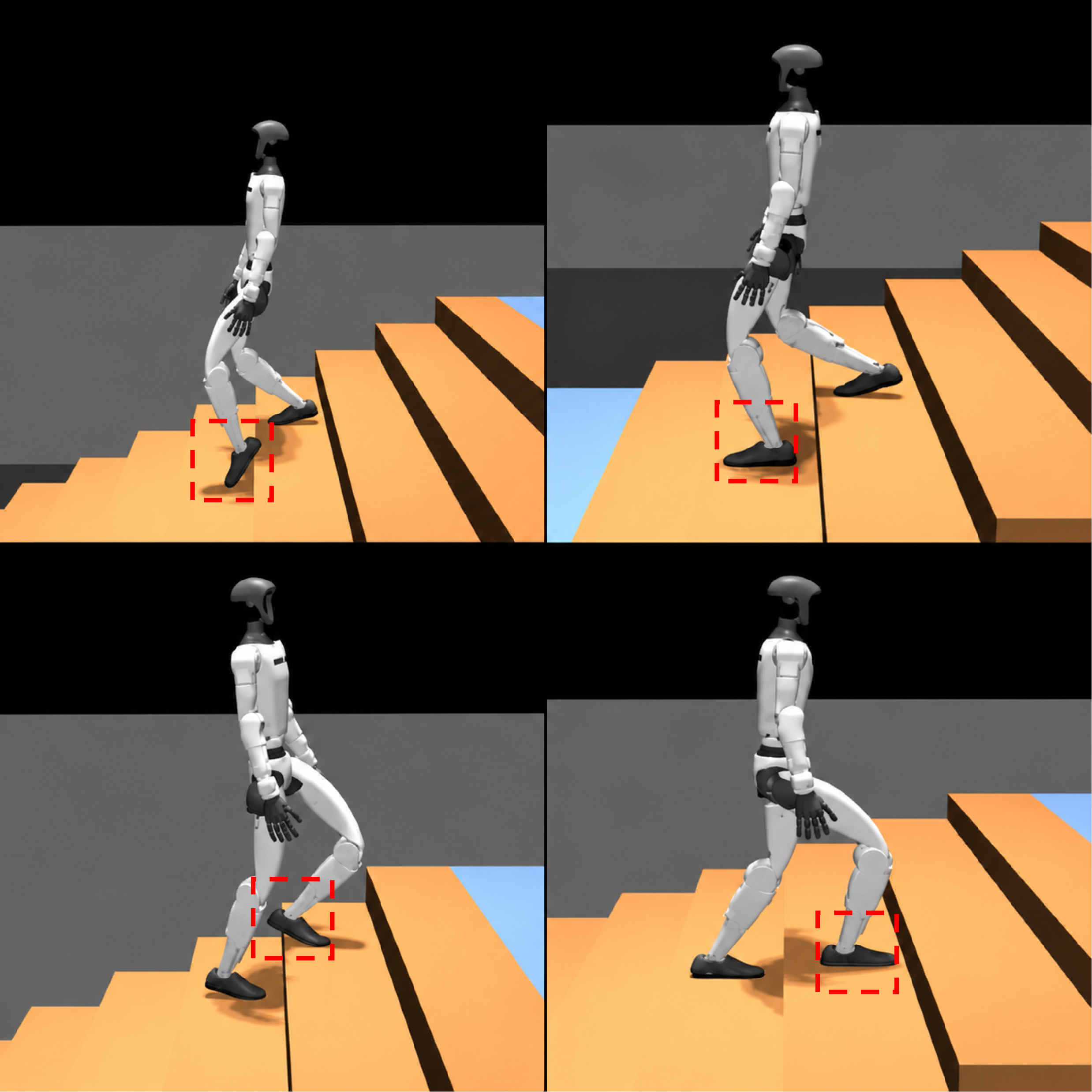}
\caption{Selected stair contacts without (left) and with (right) tread-midline shaping during ascent (top) and descent (bottom). Red dashed boxes identify the evaluated contacts.}
\label{fig:centerline-contact-comparison}
\end{figure}

The examples in Fig.~\ref{fig:centerline-contact-comparison} illustrate tread-midline shaping during ascent and descent. Without shaping, contacts lie near tread edges with partial sole support; with shaping, they lie farther inside, increasing visible support margins. These examples complement the lower NoMidline success at the largest stair setting; contact-distance, supported-area, and edge-frequency distributions remain unmeasured.

The traces and snapshots illustrate earlier optimization progress with force assistance and more interior contacts with tread-midline shaping. Component comparisons are aggregate checkpoint results with unavailable trial counts, seed variability, and checkpoint provenance.

\begin{figure*}[t]
\centering
\includegraphics[width=0.95\linewidth]{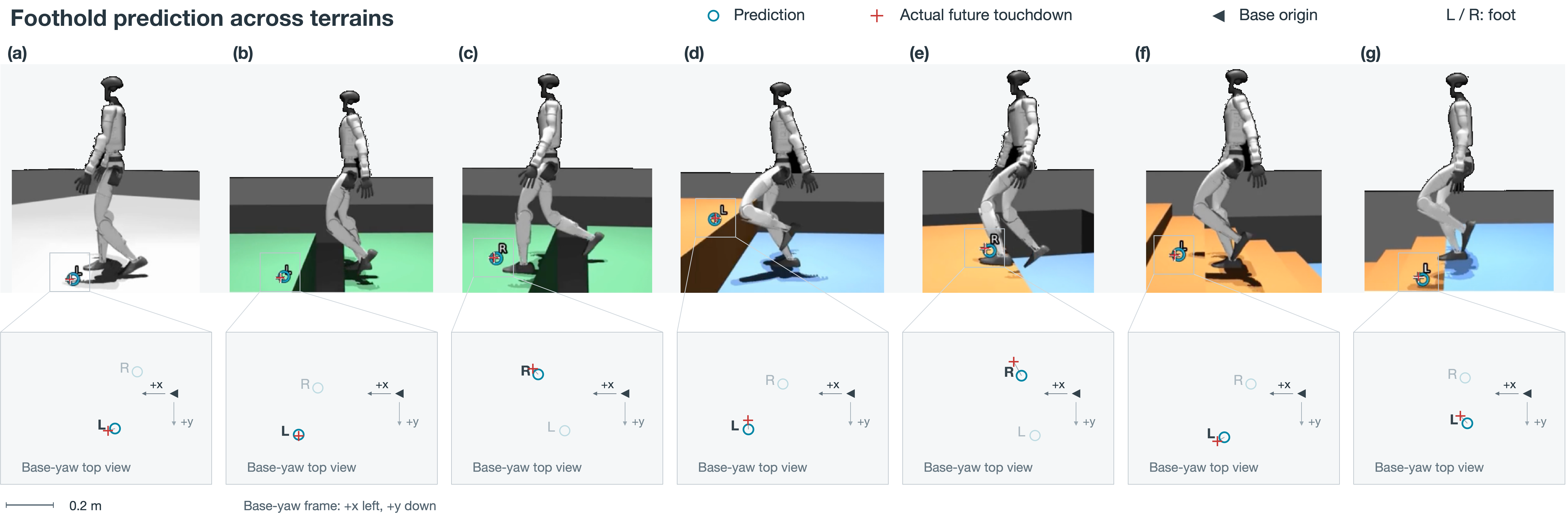}
\caption{Touchdown predictions on (a) flat ground, (b,c) gaps, (d,e) platforms, and (f,g) stairs. Top: scene views. Bottom: current base-yaw XY coordinates, with $+x$ left and $+y$ down. Cyan circles: predictions; red crosses: realized future contacts recorded offline; gray triangles: base origin. L/R identify the foot.}
\label{fig:sim-foothold-prediction}
\end{figure*}

\begin{figure*}[t]
\centering
\includegraphics[width=0.95\linewidth]{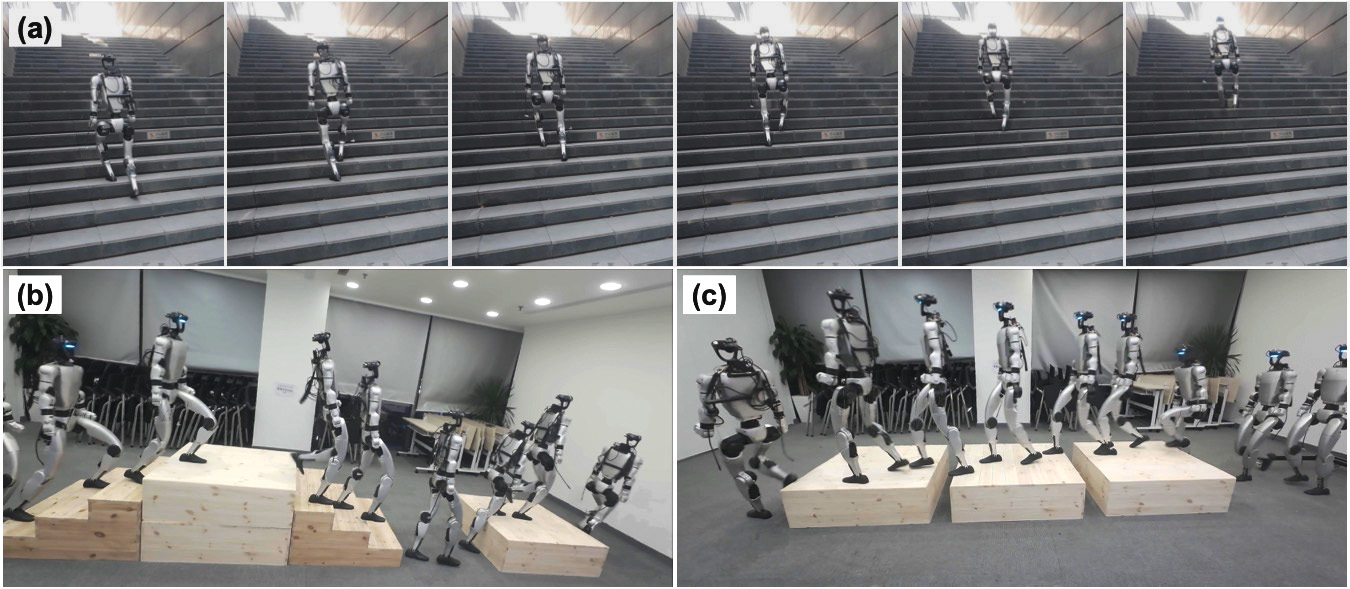}
\caption{Real-world traversal with one policy. (a) Outdoor stair ascent. (b) An indoor route combining stair ascent/descent and platforms. (c) An indoor route combining platforms and gaps. Overlaid robot poses in (b,c) show successive execution stages.}
\label{fig:real-world}
\end{figure*}

\subsection{Comparison with Released Perceptive Controllers}
\label{sec:released-comparison}
In addition to HPL, we compare \method{} with MoRE~\cite{Wan25MoRE} and Hiking~\cite{Zhu26}, using the pretrained models released with their official code (for Hiking, the stair-walking checkpoint) without retraining. All methods are evaluated in MuJoCo with identical physics settings and torque limits (Table~\ref{tab:released-comparison}). \method{} succeeds in all trials. MoRE stalls in front of the 40-cm platform until the timeout, and Hiking fails on both the platform and the gaps.

\begin{table}[t]
\centering
\caption{Success rates (\%) in simulation, 10 trials per terrain.}
\label{tab:released-comparison}
\footnotesize
\setlength{\tabcolsep}{3pt}
\begin{tabular}{@{}lccccccc@{}}
\toprule
Method & Flat & Rough & Slope & Platform & Gaps & Stairs & All\\
 & & \scriptsize 0--10\,cm & \scriptsize $20^\circ$ & \scriptsize 40\,cm & \scriptsize 45\,cm & \scriptsize 15$\times$30\,cm & \\
\midrule
MoRE & 100 & 100 & 100 & 0 & 100 & 100 & 87.5\\
Hiking & 100 & 100 & 100 & 0 & 0 & 100 & 75.0\\
Ours & 100 & 100 & 100 & \textbf{100} & 100 & 100 & \textbf{100}\\
\bottomrule
\end{tabular}
\end{table}

\subsection{Historical Visibility and Attention Grounding}
We analyze historical visibility and attention alignment during stair traversal using a policy that controls the 12 leg joints. Ascending and descending 32-cm-wide stairs at levels 0/5/9 are each recorded with eight environments for 36~s using seed 47. A sample is a prediction-time/environment/foot tuple with a legal touchdown in the complete future 24-step window and 43 real same-episode camera frames, excluding reset padding. Samples can share a touchdown; visibility measures the sole center. Among 79,537 stair samples, current-ROI visibility is 7.82\%, compared with 67.89\% in any retained frame (Table~\ref{tab:visibility-evidence}). History supplies otherwise unavailable contact evidence in 60.07\% of samples, motivating the retrieval in Fig.~\ref{fig:motivation}. The remaining 32.11\% are unseen within the retained window.

\begin{table}[t]
\centering
\caption{Future touchdown visibility and attention alignment during stair traversal.}
\label{tab:visibility-evidence}
\footnotesize
\begin{tabular}{@{}lr@{}}
\toprule
Diagnostic & Value\\
\midrule
Current-frame ROI visible & 7.82\%\\
History-only ROI visible & 60.07\%\\
Any retained ROI frame visible & 67.89\%\\
\midrule
All-head mean positive-set mass & 34.30\%\\
Uniform positive-set mass reference & 5.58\%\\
GT-best-head Top-1 hit & 81.73\%\\
GT-best-head Top-3 hit & 95.80\%\\
\bottomrule
\end{tabular}
\end{table}

On 53,994 history-visible samples, all-head mean positive-set mass is 34.30\%, versus a 5.58\% uniform reference over all 288 tokens. GT-best-head Top-1 and Top-3 hit rates are 81.73\% and 95.80\%. Set mass sums each head's attention over positive tokens and then averages across heads, without bilinear label weights. The uniform reference ignores the valid-token mask. GT-best-head selects the head with greatest ground-truth-weighted overlap before checking its top-ranked tokens. These offline metrics characterize contact alignment in the diagnostic checkpoint.

\subsection{Prediction Quality and Policy Utilization}
Fig.~\ref{fig:sim-foothold-prediction} compares predicted and actual future touchdowns across flat ground, gaps, platforms, and stairs. The planar panels expose offsets obscured by scene perspective. Predictions are close to their corresponding contacts in the displayed states, with visible discrepancies providing qualitative context for the error measurements below.

Offline replay of the diagnostic stair trajectories yields touchdown labels for 49.35\% of time-foot pairs with a complete future 24-step window, versus 21.94\% when labels must mature within the current rollout. Rollout boundaries reduce supervision availability. Per-coordinate XY MAE rises from 1.31~cm at a three-step lead to 8.52~cm at a 24-step lead, indicating greater prediction error farther ahead of contact.

NoFootQuery removes multiple paths, so its traversal results and the offline alignment diagnostic leave the actor's dependence on contact-specific retrieval unresolved. Matched no-read-supervision and closed-loop context-zero/context-shuffle controls would separate this dependence from global features, frame-lag preferences, and common image locations.

\subsection{Real-World Experiments}
\label{sec:real-world}

We deploy \method{} on a Unitree G1 humanoid using the same policy across outdoor stairs and indoor courses combining stair ascent and descent, platform traversal, and gap crossing.

In Fig.~\ref{fig:real-world}(a), the robot ascends outdoor stairs by alternating support between adjacent treads while maintaining balance.

The indoor routes test transitions between terrain types. The robot links stair ascent, descent, and platform traversal in Fig.~\ref{fig:real-world}(b), steps onto platforms, crosses intervening gaps, and steps down in the route of Fig.~\ref{fig:real-world}(c). These connected motions show continuous execution across changes in support height and surface continuity.

\section{Discussion and Conclusion}
\method{} uses predicted contacts to organize depth-history retrieval, with realized-contact visibility providing direct supervision. Global memory complements per-foot reads, while force assistance and tread-midline shaping support training. Simulation comparisons favor the complete system at the largest tested obstacles. The separate diagnostic quantifies historical visibility, read alignment, and increasing prediction error with lead time. A single policy executes mixed-terrain motion in simulation and on the Unitree G1, including outdoor stair ascent and indoor stair, platform, and gap traversal, as shown in Fig.~\ref{fig:real-world}.

\section*{Acknowledgment}
We used Cursor and ZCode to assist with manuscript preparation and language editing.

\begingroup\footnotesize
\bibliographystyle{IEEEtran}
\bibliography{references}
\endgroup

\end{document}